%% file: main.tex
\documentclass[runningheads]{llncs}

\usepackage{eccv}

\usepackage{eccvabbrv}

\usepackage{graphicx}
\usepackage{booktabs}
\usepackage{tabularx}
\usepackage{multirow}
\usepackage{makecell}
\usepackage{hhline}
\usepackage{mathrsfs}
\usepackage{dsfont}
\usepackage{pifont}

\usepackage[accsupp]{axessibility}  % Improves PDF readability for those with disabilities.

\newcommand{\cmark}{\ding{51}}
\newcommand{\xmark}{\ding{55}}

\usepackage[table]{xcolor}
\definecolor{bestred}{RGB}{255,120,120}
\definecolor{secondorange}{RGB}{255,190,100}
\definecolor{thirdyellow}{RGB}{255,240,100}

\usepackage{orcidlink}

\usepackage{hyperref}
\usepackage[protrusion=true,expansion=false]{microtype}
\begin{document}

% ---------------------------------------------------------------
\title{Fast and Compact 3D Gaussian Splatting with Polarized Opacity Prior}

\titlerunning{Fast and Compact 3DGS with Polarized Opacity Prior}

\author{Zi-Ming Wang\inst{1, 2}\orcidlink{0009-0003-4644-2349} \and Kai-Wen Duan\inst{1}\orcidlink{0009-0007-4348-2694} \and Kowei Huang\inst{1}\orcidlink{0009-0005-6057-2697} \and Akihiro Sugimoto\inst{2}\orcidlink{0000-0001-9148-9822} \and Shang-Hong Lai\inst{1}\orcidlink{0000-0002-5092-993X}}

\authorrunning{Z.~Wang, K.~Duan et al.}

\institute{National Tsing Hua University, Taiwan \\
\email{\{ziming614, kevin77688, st114065531\}@gapp.nthu.edu.tw, lai@cs.nthu.edu.tw} \and National Institute of Informatics, Japan \\
\email{sugimoto@nii.ac.jp}}

\maketitle

% % ===================================================================
% % Teaser
% % ===================================================================
\begin{figure}[tb]
    \centering
    \includegraphics[width=0.85\linewidth]{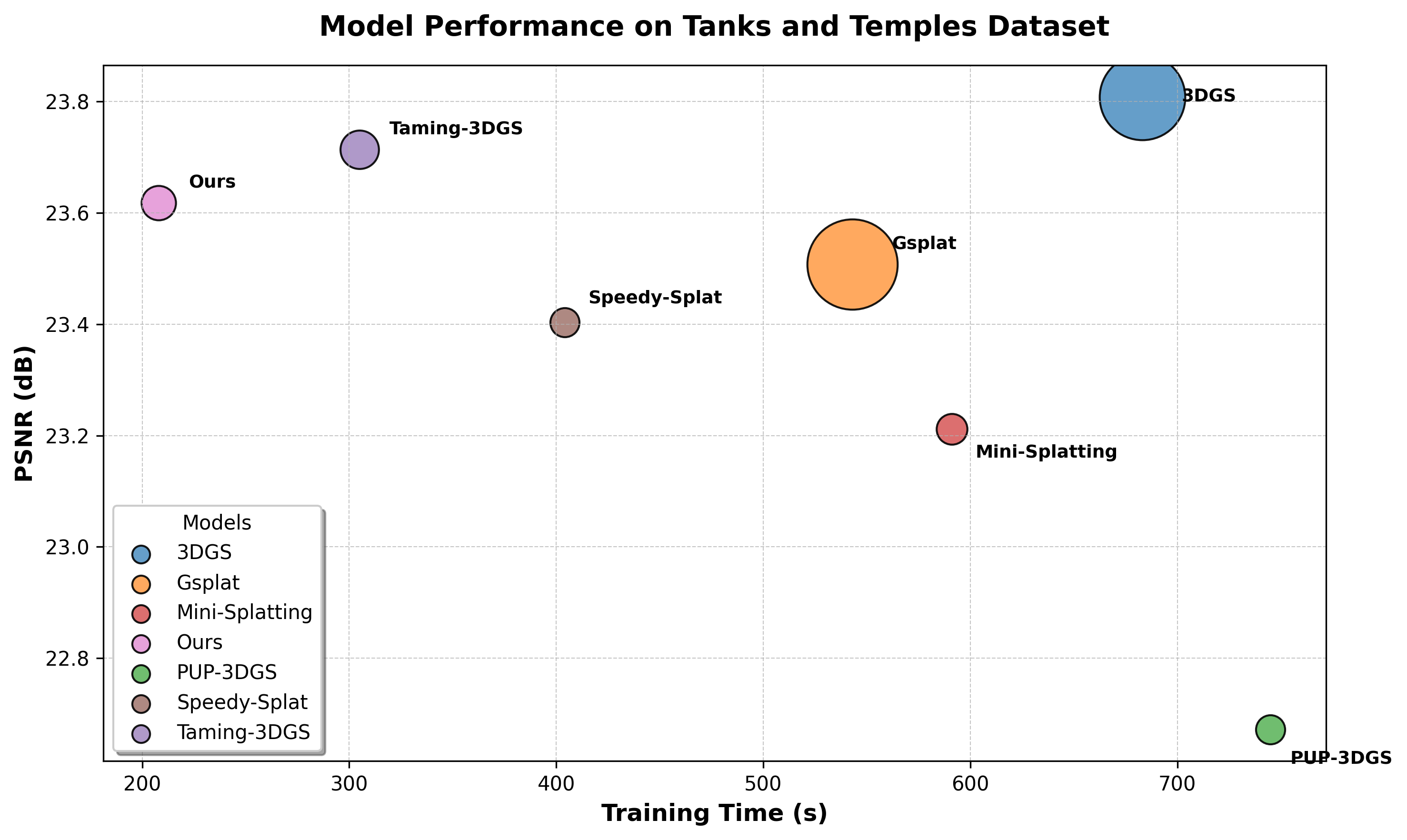}
    \caption{Average PSNR versus training time on Tanks and Temples dataset~\cite{tanks}. Our method achieves the fastest training speed among state-of-the-art approaches while maintaining competitive reconstruction quality. The size of each marker indicates the number of Gaussians used in the corresponding model.}
    \label{fig:teaser}
\end{figure}

% % ===================================================================
% % ABSTRACT
% % ===================================================================
\begin{abstract}
3D Gaussian Splatting (3DGS) achieves state-of-the-art rendering quality at real-time speeds but suffers from ``model bloat''---a large number of redundant, low-opacity Gaussians that inflate memory usage and training costs. This inefficiency stems from the standard ``densify-then-prune'' paradigm, which expands the model aggressively before relying on pruning to achieve compactness.
To mitigate this problem, we present an efficient training framework that builds an \textbf{intrinsically compact} representation, replacing the conventional \emph{densify-then-prune} cycle. Our method leverages a synergistic design: an $\mathcal{L}_{2}$ reconstruction loss to provide \textbf{error-proportional gradients} that stabilize optimization, and a novel \textbf{Polarized Opacity Prior (POP)} to actively manage the Gaussian population. POP steers informative primitives toward full opacity and uninformative ones toward transparency, enabling natural pruning and accelerating rendering through Early Ray Termination.
Experiments on three public datasets %(Mip-NeRF 360, Tanks \& Temples, and Deep Blending) 
demonstrate that our approach consistently achieves accelerated 3DGS training with significantly fewer Gaussians while maintaining comparable visual reconstruction quality. These results show that the proposed framework provides a simple and effective path toward fast and inherently compact 3DGS training.

\keywords{3D Gaussian Splatting \and Polarized Opacity Prior \and $\mathcal{L}_{2}$ Loss \and Compact Representation \and Efficient Training}
\end{abstract}

% ===================================================================
% SECTION 1: Introduction
% ===================================================================
\section{Introduction}
\label{sec:intro}

3D Gaussian Splatting (3DGS)~\cite{3DGS} has established a new standard for high-quality, real-time novel view synthesis. However, its practical deployment is severely hindered by ``model bloat'' resulting from the conventional ``densify-then-prune'' paradigm. This strategy typically generates millions of redundant, low-opacity Gaussians, consuming excessive memory and wasting computational resources on primitives that contribute negligibly to the final scene. Unlike most existing compression techniques that function as post-processing steps, we argue that the fundamental inefficiency lies in the training process itself---specifically, in the aggressive expansion of primitives and the reliance on heuristic ``opacity resets'' for pruning.

To address these challenges, we propose an efficient training framework that maintains a compact representation throughout the optimization process. Our core insight is that the stability of Gaussian evolution is tied to the relationship between reconstruction loss and opacity management. We introduce a synergistic mechanism that couples stable gradient feedback with a \textbf{Polarized Opacity Prior (POP)}, which significantly reduces Gaussian counts and accelerates convergence. The efficacy of this synergy is showcased in \cref{fig:teaser}, where our approach establishes a new efficiency benchmark—attaining competitive PSNR with a drastically reduced memory footprint and shorter training times compared to current state-of-the-art methods. Specifically, our work makes the following contributions:

\begin{itemize}
    \item \textbf{Analysis of the Gradient-Regularization Trade-off:} We identify that standard $\mathcal{L}_{1}$ loss provides insufficient gradient magnitude to counteract strong opacity regularization, leading to artifacts. We propose using an $\mathcal{L}_{2}$ formulation to provide error-proportional gradients, which stabilize optimization under sparsity constraints.
    \item \textbf{Polarized Opacity Prior (POP):} We introduce a novel regularization term that steers informative primitives toward full opacity while driving redundant ones toward transparency. This polarization encourages Early Ray Termination, thereby accelerating rendering.
    \item \textbf{Efficient Training Framework:} We develop a framework that replaces the original opacity-reset cycle with POP, enabling natural and continuous pruning. This approach achieves state-of-the-art training speeds and produces significantly smaller models while delivering rendering quality comparable to current state-of-the-art methods. 
\end{itemize}

% ===================================================================
% SECTION 2: Related Work
% ===================================================================
\section{Related Work}
\label{sec:related_work}

\subsection{Novel View Synthesis and 3DGS}
\label{subsec:nvs_3dgs}
Neural Radiance Fields (NeRF)~\cite{NeRF} pioneered neural implicit functions for photorealistic synthesis, yet heavy MLP evaluations limit real-time use. Subsequent research accelerated this via explicit or hybrid structures, such as voxel grids~\cite{DVGO, DVGO_v2}, MLP factorization~\cite{FastNeRF}, and multiresolution hash encoding~\cite{instantNGP}. Notably, TensoRF~\cite{TensoRF} utilizes low-rank tensor factorization to reduce memory footprint. A major breakthrough, 3D Gaussian Splatting (3DGS)~\cite{3DGS}, employs explicit primitives and tile-based rasterization for real-time rendering. Despite its efficiency, the Gaussian count often grows explosively, leading to significant memory overhead~\cite{Recent_Advances_in_3DGS, A_Survey_on_3DGS}. This scalability bottleneck has catalyzed research into restructuring densification to make 3DGS more compact and training-efficient.

\subsection{Compact Representations for 3DGS}
\label{subsec:compression}
Existing methods generally focus on reducing redundancy through pruning and quantization, either as a post-optimization step applied after full model expansion or during joint training.

Post-processing methods utilize significance scoring or vector quantization (VQ) to compress pre-trained models. LightGaussian~\cite{lightGS} employs a Global Significance Score based on volume and opacity, alongside knowledge distillation for spherical harmonics, to achieve over $15\times$ compression. Compressed-3DGS~\cite{Compressed3DGS} utilizes a sensitivity-aware vector clustering approach based on image energy contribution and linearizes Gaussians along a Morton-order space-filling curve to maximize spatial coherence, while employing quantization-aware fine-tuning alongside the clustering process, resulting in a $31\times$ compression ratio. PUP-3DGS~\cite{pup3dgs} introduces a multi-round prune-refine pipeline guided by a mathematically principled sensitivity pruning score. Derived from a second-order approximation of the reconstruction error, the score utilizes a block-wise Fisher information matrix centered on spatial parameters to quantify geometric uncertainty, thereby removing up to 90\% of Gaussians with minimal quality loss. 

Joint compression approaches~\cite{CompGS, compact3DGS_KT, HAC} integrate quantization directly into the optimization loop. Methods like CompGS~\cite{CompGS} utilize K-means-based quantization on centroids, while Compact-3DGS~\cite{compact3DGS_KT} replaces high-dimensional spherical harmonics with grid-based neural fields, incorporates residual vector quantization (R-VQ)~\cite{r-vq} for geometric attributes, and utilizes learnable masking to dynamically prune non-essential Gaussians. Other approaches, such as Self-Organizing Gaussian Grid~\cite{compact3dgs_grid}, periodically re-sort parameters into 2D grids using parallel linear assignment sorting, yielding reduction factors of 17--42$\times$. Finally, Mini-Splatting~\cite{miniSplatting} focuses on spatial reorganization. It restructures densification through blur-based splitting and strictly constrains the primitive budget using importance-weighted sampling, thereby maintaining high-quality reconstruction with minimal primitives.

Unlike these memory-intensive \textit{``densify-then-prune''} approaches that suffer from early Gaussian proliferation, our continuous optimization objective naturally restricts primitive growth from the outset. This ensures a stable, compact representation while maintaining comparable rendering quality.

\subsection{Efficient Training of 3DGS}
\label{subsec:fast_training}
Another line of work focuses on reducing training latency by addressing bottlenecks ranging from low-level kernel optimizations to high-level geometric pruning. At the hardware level, DISTWAR~\cite{DISTWAR} identifies that atomic operations create L2 cache contention and proposes a novel primitive for warp-level reduction, achieving a $2.44\times$ speedup in gradient computation and $1.41\times$ for the entire training pipeline. Moving to system-algorithm co-design, LiteGS~\cite{LiteGS} optimizes CUDA kernels using a "Cluster-Cull-Compact" pipeline that enhances spatial locality and L2 cache hit rates via Morton-code reordering, and implements warp-based rasterization. Taming-3DGS~\cite{Taming_3DGS} targets the backpropagation bottleneck with per-splat parallelization to reduce atomic collisions alongside batched attribute updates, and replaces heuristic densification with score-based, budget-controlled sampling of Gaussian primitives. Speedy-Splat~\cite{Speedy-Splat} integrates soft and hard pruning directly into the training loop, which is coupled with SnugBox and AccuTile for precise rendering, thereby discarding over 80\% of unnecessary Gaussians to achieve a $1.47\times$ faster training speed. 

In contrast to the aforementioned methods, our framework accelerates training by processing fewer primitives. Through early suppression of uncontrolled Gaussian proliferation, we significantly reduce training overhead with only minimal modifications to low-level kernels.

\subsection{Opacity Entropy Loss in 3DGS}
\label{subsec:opacity_entropy_loss}
Recent works incorporate an opacity entropy loss to binarize opacities ($\alpha \in \{0,1\}$), driving scene primitives toward definitive physical states to precisely delineate solid 3D surface boundaries.

For instance, SuGaR~\cite{sugar} extracts precise polygonal meshes by aligning Gaussians with true object surfaces via SDF-based regularization. Because this formulation assumes perfectly flat and opaque primitives, an opacity entropy loss is applied to penalize intermediate values and prune semi-transparent volumes. This structural prior bypasses the sparsity issues of traditional Marching Cubes in favor of efficient depth-map sampling and Poisson reconstruction for fast and accurate meshing. Similarly, HaloGS~\cite{halogs} addresses geometric redundancy via a dual-representation framework that loosely couples scene geometry and appearance. It fits surface geometry with learnable triangles that decode neural Gaussians for high-frequency appearance. To ensure compactness, HaloGS applies an opacity entropy loss during fine-tuning to binarize triangle opacities, periodically pruning primitives with low opacity or minimal view-space contribution. This structurally regularized foundation enables the efficient extraction of Level-of-Detail (LoD) planar primitives and compact meshes.

Ultimately, while both approaches demonstrate that enforcing binary opacity serves as a crucial regularizer for superior surface extraction, our work introduces a novel perspective, leveraging opacity binarization specifically to address the gradient leakage problem (\cref{subsec:motivation}) during training.

% ===================================================================
% SECTION 3: Methodology
% ===================================================================
\section{Methodology}
\label{sec:methodology}

\subsection{Preliminary: 3D Gaussian Splatting}
\label{subsec:preliminary}
3DGS~\cite{3DGS} represents a scene as a set of anisotropic Gaussians $\{\mathcal{G}_i\}$. Each Gaussian is parameterized by its position $\boldsymbol{\mu}_i$, color $\mathbf{c}_i$, opacity $o_i \in [0, 1]$, and a covariance matrix $\boldsymbol{\Sigma}_i$ (factorized into a scale vector $\mathbf{s}_i$ and rotation quaternion $\mathbf{q}_i$). During rendering, these 3D Gaussians are projected onto the 2D image plane through a local affine transformation~\cite{EWA_Splatting}. The final color of a pixel $\mathbf{p}$ is computed via alpha blending:
\begin{equation}\label{eq:render}
    \mathbf{c}(\mathbf{p}) = \sum_{i=1}^{m} T_i \, \alpha_i \, \mathbf{c}_i, \quad \alpha_i = o_i \, \mathcal{G}_i^{2D}(\mathbf{p}).
\end{equation}
where $\alpha_i$ is the per-pixel opacity of $\mathcal{G}_i$, and $T_i = \prod_{j=1}^{i-1} (1 - \alpha_j)$ denotes the accumulated transmittance. Here, $T_i$ represents the remaining blending weight available for the $i$-th Gaussian; as more objects are encountered along the ray, $T_i$ decreases. To optimize efficiency, rendering triggers \textbf{Early Ray Termination} if the remaining weight becomes negligible, i.e., $T_{i+1} \leq 10^{-4}$. The standard training objective minimizes a weighted combination of $\mathcal{L}_1$ and SSIM loss:
\begin{equation}\label{eq:3DGS_color_loss}
    \mathcal{L}_{\text{color}} = (1 - \lambda_{\text{SSIM}}) \mathcal{L}_{1} + \lambda_{\text{SSIM}} \mathcal{L}_{\text{D-SSIM}}.
\end{equation}

\paragraph{Adaptive Density Control.} 3DGS is typically initialized from a sparse point cloud obtained via Structure-from-Motion (SfM)~\cite{sfm}. To accurately capture scene geometry, 3DGS periodically performs densification and pruning (typically every 100 iterations). Gaussians are selected for densification if their average view-space gradient magnitude exceeds a threshold $\tau_{\mathrm{Den}}$, as formulated in \cref{eq:densification}, where $k$ denotes the index of the training view and $\mathbf{V}$ represents the densification period (i.e., 100). Moreover, small Gaussians undergo \textit{cloning}, while large Gaussians are \textit{split} into smaller primitives.

\begin{equation} \label{eq:densification}
    \frac{1}{\mathbf{V}} 
    \sum_{k=1}^{\mathbf{V}}
    \sqrt{
        \left( 
        \frac{\partial \mathcal{L}_k}{\partial \mu_{i}^{2D,x}}
        \right)^2
        +
        \left(
        \frac{\partial \mathcal{L}_k}{\partial \mu_{i}^{2D,y}}
        \right)^2
    }
    > \tau_{\mathrm{Den}}.
\end{equation}

On the other hand, redundant or poorly positioned Gaussians are \textit{pruned} if their opacity $o_i$ falls below a certain threshold or if they grow excessively large. A critical heuristic is the \textbf{opacity reset}—periodically setting all $o_i \to 0$ every 3,000 iterations—to force the optimization to re-justify the importance of each Gaussian, facilitating the removal of redundant primitives.

\subsection{Motivation}
\label{subsec:motivation}

\begin{figure}[tb]
    \centering
    \includegraphics[width=0.5\linewidth]{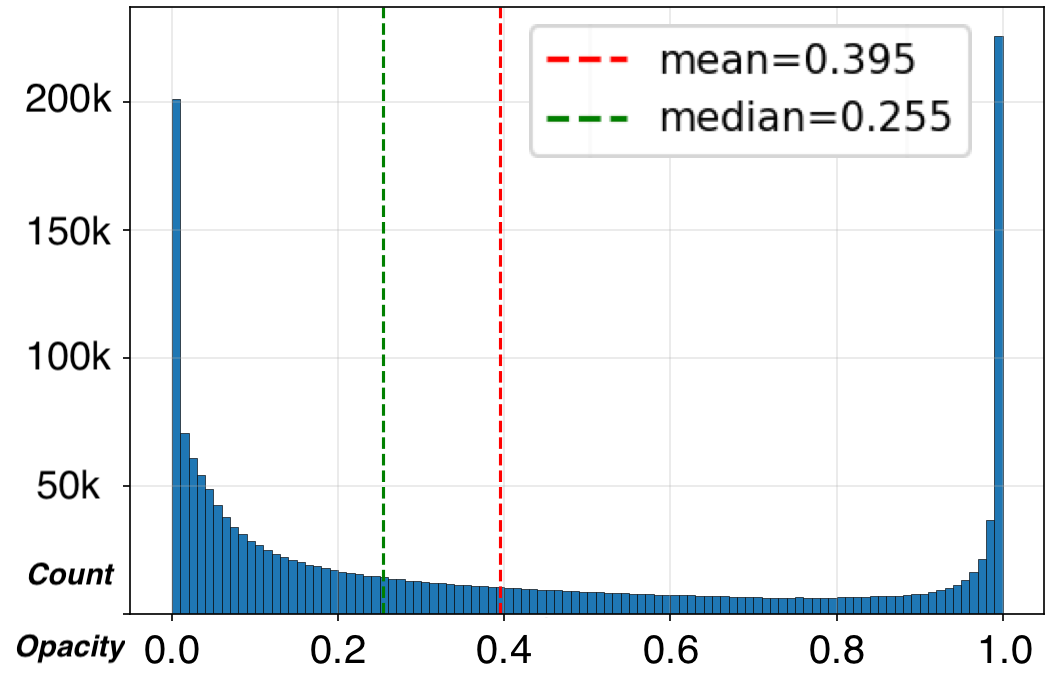}
    \caption{Opacity statistics on the Kitchen scene~\cite{Mip_NeRF_360}. Approximately 19.9\% of Gaussians have opacity $> 0.9$, while 33.7\% remain at opacity $< 0.1$, indicating a large number of redundant primitives.}
    \label{fig:3DGS_opacity}
\end{figure}

\begin{figure}[tb]
    \centering
    \includegraphics[width=0.9\linewidth]{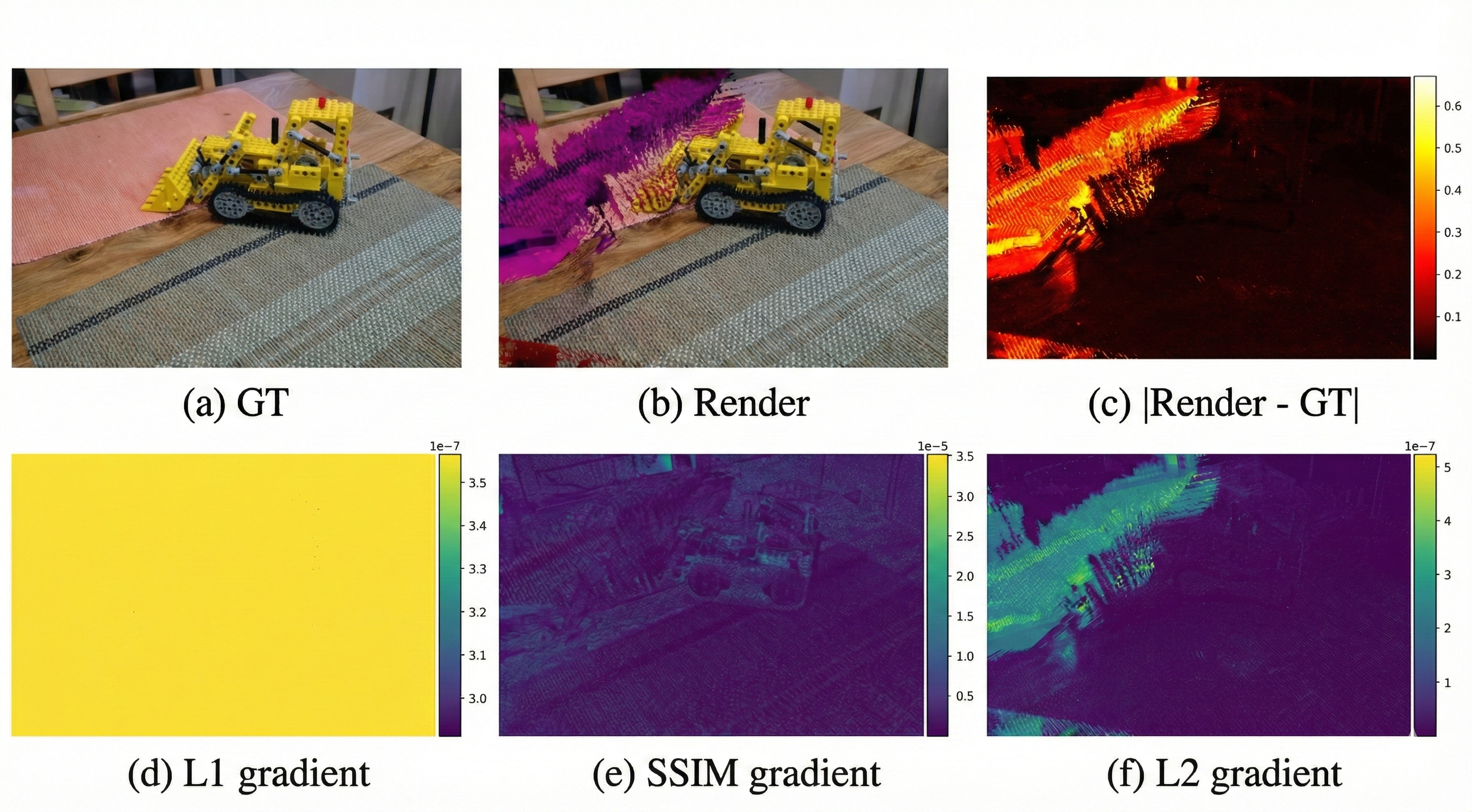}
    \caption{Artifact demonstration on the Kitchen scene when forcing $o_i \to 1$ for all Gaussians: (a) Ground Truth, (b) Rendered image, (c) Error map $|Render - GT|$, (d) $\mathcal{L}_1$ loss gradient w.r.t. pixel color, (e) SSIM loss gradient w.r.t. pixel color, (f) $\mathcal{L}_2$ loss gradient w.r.t. pixel color.}
    \label{fig:3DGS_Opacity=1_artifact}
\end{figure}

Our approach is motivated by a critical observation: \textit{Why do numerous redundant, low-opacity Gaussians survive the pruning process despite contributing negligibly to the final image?} (see \cref{fig:3DGS_opacity}). 

We identify this as \textbf{Gradient Leakage}: In the standard 3DGS pipeline, even Gaussians with near-zero opacity receive persistent gradient updates accumulated from a vast number of pixels. This collective gradient often suffices to counteract the opacity reset mechanism, allowing redundant primitives to ``drift'' back into existence without ever providing significant geometric value. 

Our hypothesis is that by driving important Gaussians toward full opacity ($o_i \to 1$), we can effectively trigger Early Ray Termination. When the accumulated transmittance $T_i$ drops below a threshold rapidly, the rendering engine skips subsequent Gaussians in the depth order. This not only accelerates the forward pass but also halts backpropagation to redundant primitives. This creates a virtuous cycle: fewer, high-opacity Gaussians represent the scene, leading to a more compact model and faster convergence.

However, naive opacity regularization (i.e., forcing $o_i \to 1$ for all) introduces severe visual artifacts. As shown in \cref{fig:3DGS_Opacity=1_artifact} (panel b), we observe that the standard $\mathcal{L}_1 + \text{SSIM}$ loss combination may fail to correct these artifacts. This is because neither the gradient magnitude of $\mathcal{L}_1$ nor SSIM (\cref{fig:3DGS_Opacity=1_artifact}, panels d and e) has a direct relationship with the reconstruction error (\cref{fig:3DGS_Opacity=1_artifact}, panel c), leading to a fundamental optimization conflict with the regularization term.

\subsection{$\mathcal{L}_2$ Loss and Blur-Based Densification}
\label{subsec:l2_loss}
Standard 3DGS employs a combination of $\mathcal{L}_1$ and SSIM loss for scene reconstruction. However, a significant drawback of the $\mathcal{L}_1$ loss is its ``error-agnostic'' nature: its gradient magnitude remains constant regardless of the actual reconstruction error (\cref{fig:3DGS_Opacity=1_artifact}, panel d). To address this and better suppress artifacts during optimization, we replace the primary loss with an $\mathcal{L}_2$ formulation. For a pixel $\mathbf{p}$ in an image of resolution $H \times W$, the gradients of $\mathcal{L}_1$ and $\mathcal{L}_2$ loss are:
\begin{equation}\label{eq:loss_grads}
    \frac{\partial \mathcal{L}_{1}}{\partial \mathbf{c}(\mathbf{p})} = \frac{\operatorname{sign}(\Delta \mathbf{c})}{3HW}, \quad \frac{\partial \mathcal{L}_{2}}{\partial \mathbf{c}(\mathbf{p})} = \frac{2 \Delta \mathbf{c}}{3HW}.
\end{equation}
where $\Delta \mathbf{c} = \mathbf{c}(\mathbf{p}) - \mathbf{c}^{\mathrm{GT}}(\mathbf{p})$. 

\begin{figure}[tb] 
    \centering
    % 第一張子圖 (a)
    \begin{subfigure}{0.48\linewidth}
        \centering
        \includegraphics[width=\linewidth]{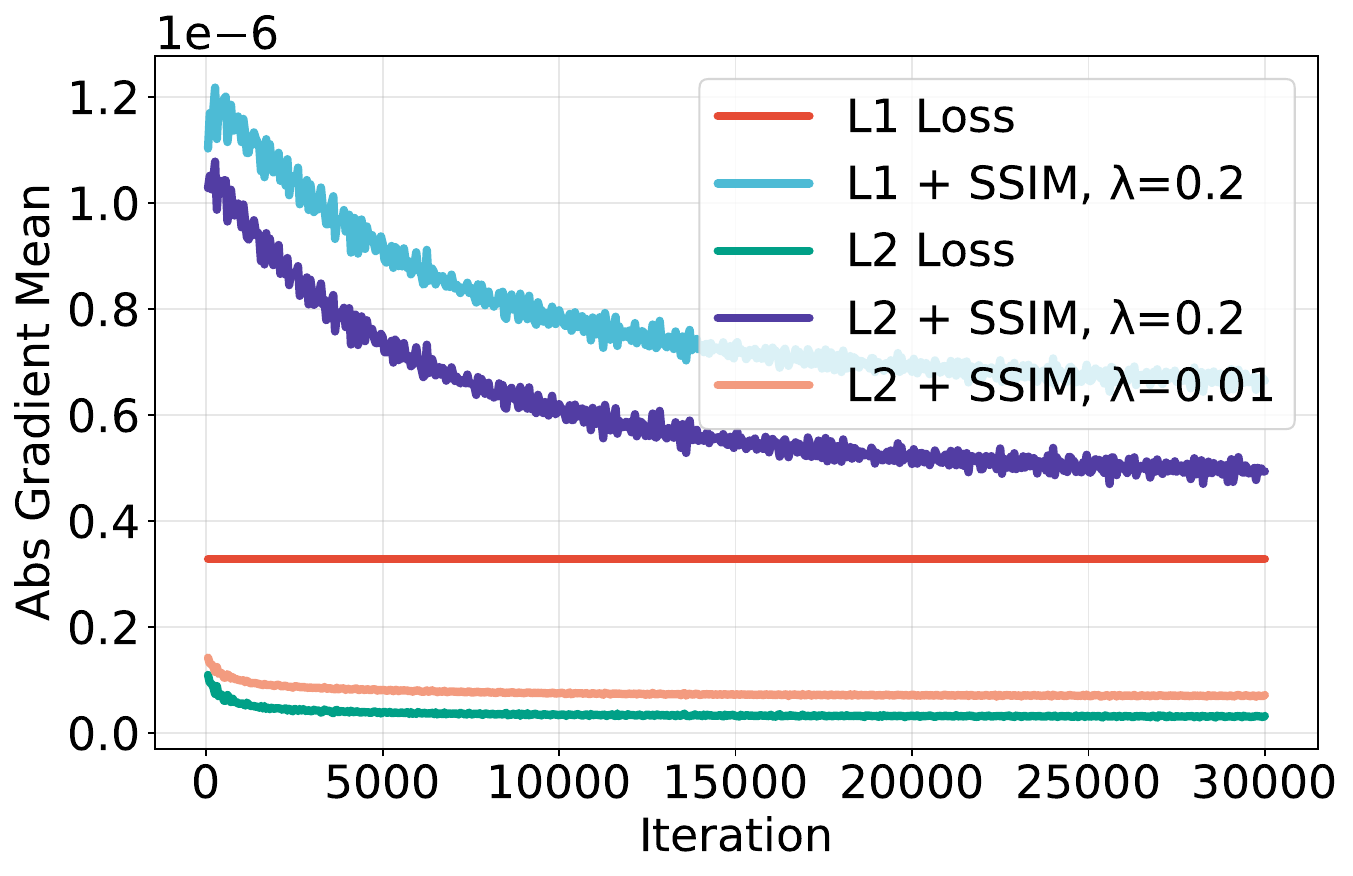}
        % 確實移除了句號，並精簡字數
        \caption{Comparison of gradient magnitudes $\left| \frac{\partial \mathcal{L}}{\partial \mathbf{c}} \right|$ on the Bicycle scene}
        \label{fig:grad_l1vsl2}
    \end{subfigure}
    \hfill % 自動填滿中間的間隔
    % 第二張子圖 (b)
    \begin{subfigure}{0.48\linewidth}
        \centering
        \includegraphics[width=\linewidth]{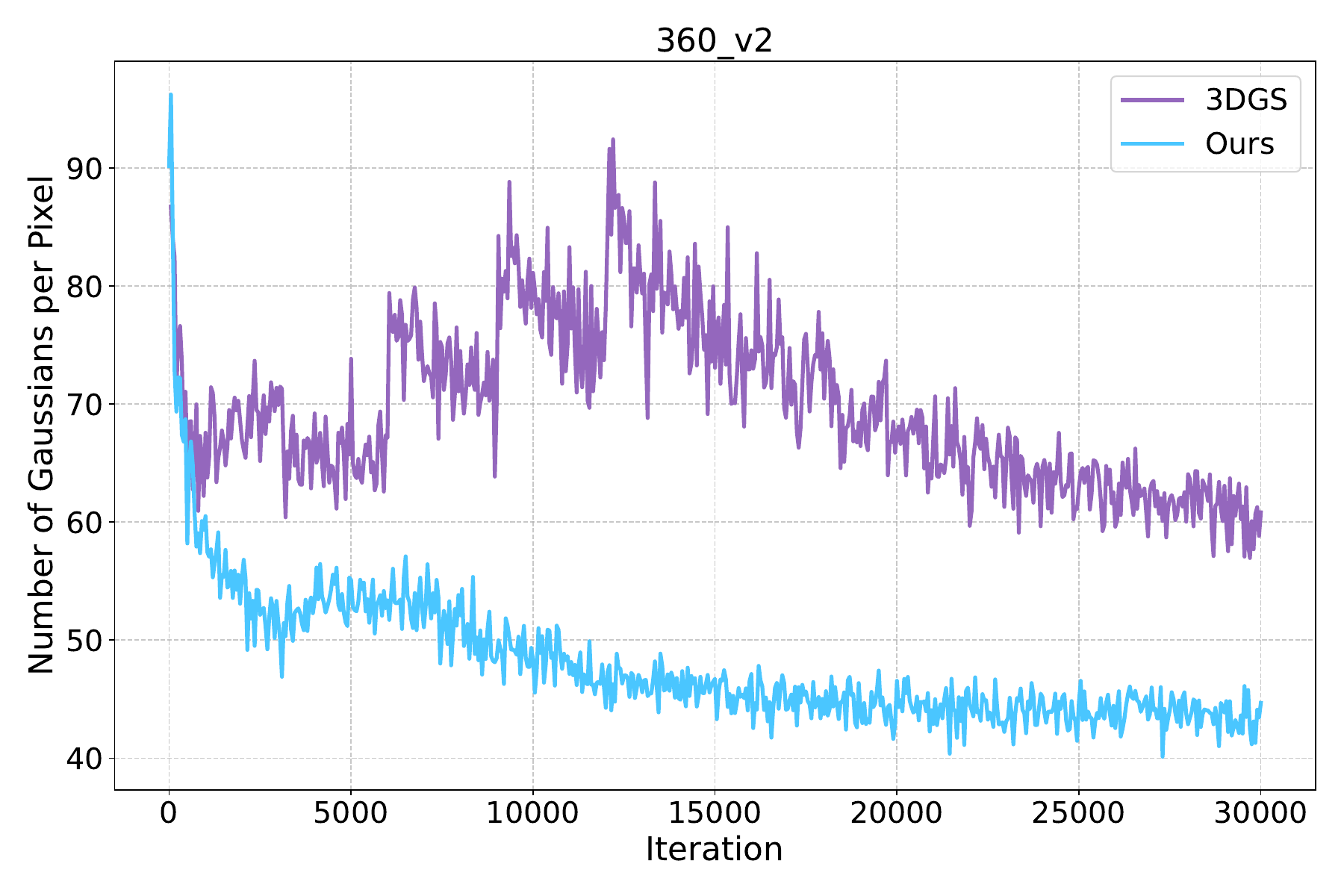}
        \caption{Average Gaussians per pixel across iterations}
        \label{fig:perPixel_GSs_360_v2}
    \end{subfigure}
    
    % 整張大圖的總標題
    \caption{Effectiveness of our proposed method on Mip-NeRF 360~\cite{Mip_NeRF_360}. (a) $\mathcal{L}_2$ gradients (green) adapt to the error magnitude, whereas $\mathcal{L}_1$ gradients (red) remain constant. (b) POP significantly reduces the average Gaussians per pixel, enabling Early Ray Termination compared to the 3DGS baseline.}
    \label{fig:ablation_gradients_and_pop}
\end{figure}

As shown in \cref{eq:loss_grads} and \cref{fig:3DGS_Opacity=1_artifact} (panel f), the $\mathcal{L}_2$ gradient magnitude scales linearly with the color difference $\Delta \mathbf{c}$~\cite{Micro-splatting}. This allows the optimization to prioritize regions with high reconstruction error, effectively stabilizing the model when driving Gaussians toward higher opacity. Depending on the error magnitude, this gradient can range from approximately $2/255$ to $2$ times that of the $\mathcal{L}_1$ loss. In fact, as illustrated in \cref{fig:grad_l1vsl2}, the magnitude of the $\mathcal{L}_2$ gradient is typically smaller than that of the $\mathcal{L}_1$ loss under practical training conditions.

\paragraph{Blur-based Densification.}
The inherently small magnitude of $\mathcal{L}_2$ gradients poses a challenge for standard 3DGS densification, which relies on gradient-based thresholds to trigger Gaussian splitting or cloning. This limitation directly hinders the creation of new Gaussians, making the under-reconstruction problem more pronounced.

To address this, we implement a non-gradient-based densification mechanism inspired by Mini-Splatting~\cite{miniSplatting}. Their work observes that traditional gradient-based strategies often fail in regions with smooth color transitions (i.e., \textbf{blurry areas}). We introduce a \textbf{dominant count} $\mathcal{D}_i$ to identify whether a Gaussian exerts sufficient influence over the rendered image. Specifically, during rendering, we identify the Gaussian $\mathcal{G}_i$ that provides the maximum contribution to each pixel $\mathbf{p}$:

\begin{equation}\label{eq:dominant_id}
    i^*(\mathbf{p}) = \arg\max_{i} (T_i \alpha_i).
\end{equation}

For each training view, we accumulate $\mathcal{D}_i$ for every Gaussian by counting how many pixels it dominates. A Gaussian is then selected for densification---forming the set $\mathcal{G}_{\text{blur}}$ in \cref{eq:blur_split}---if its cumulative count exceeds a resolution-dependent threshold $\tau_{\text{blur}}$:

\begin{equation}\label{eq:blur_split}
    \mathcal{G}_{\text{blur}} = 
    \big\{\, \mathcal{G}_i \mid 
    \mathcal{D}_i > \tau_{\text{blur}}, \;
    \tau_{\text{blur}} = \theta_{\text{blur}} \cdot H \cdot W 
    \big\}.
\end{equation}

Since the $\mathcal{L}_2$ loss yields smaller gradients compared to the $\mathcal{L}_1$ loss used in Mini-Splatting~\cite{miniSplatting}, we set $\theta_{\text{blur}} = 2 \times 10^{-5}$, which is one-tenth of the threshold used in their work. This adjustment allows a larger population of Gaussians---specifically the smaller ones---to participate in the densification process. Furthermore, to avoid the excessive microscopic proliferation inherent in Mini-Splatting's split-only strategy, we adaptively choose between cloning and splitting based on scale. This prevents the accumulation of uninformative primitives, ensuring a compact representation while preserving fine geometric details.

\subsection{Polarized Opacity Prior (POP)}
\label{subsec:pop}

As demonstrated in \cref{subsec:motivation}, a naive, global regularization that forces $o_i \to 1$ for all Gaussians introduces severe visual artifacts. To reconcile rendering efficiency with reconstruction quality, we propose the \textbf{Polarized Opacity Prior (POP)}. Our key insight is to selectively ``polarize'' the opacity distribution: driving important Gaussians toward $o_i \approx 1$ to trigger Early Ray Termination, while simultaneously suppressing uninformative ones toward $o_i \approx 0$ to mitigate gradient leakage and eliminate redundancy.

\paragraph{Formulation.} 
We utilize the \textit{dominant count} $\mathcal{D}_i$ (defined in \cref{subsec:l2_loss}) to identify essential primitives. A binary importance mask $M_{\text{imp},i}$ is defined for each Gaussian $\mathcal{G}_i$ in a given view:

\begin{equation}\label{eq:mask_M}
    M_{\text{imp},i} = \mathds{1}[\mathcal{D}_i > 0].
\end{equation}

Specifically, a Gaussian is classified as ``important'' if it serves as the primary contributor to at least one pixel in the current view. By exclusively targeting these Gaussians, we ensure the $o_i \to 1$ objective is applied only where it is geometrically and photometrically justified. The POP loss is formulated as a two-way polarization mechanism:
\begin{equation}\label{eq:opacity_loss_all}
    \mathcal{L}_{\text{POP}} = \frac{1}{N_{\text{imp}}} \sum_{i=1}^{N} (1 - o_i) M_{\text{imp},i} + \frac{1}{N - N_{\text{imp}}} \sum_{i=1}^{N} o_i (1 - M_{\text{imp},i}).
\end{equation}
where $N$ and $N_{\text{imp}}$ denote the total and important Gaussian counts ($N_{\text{imp}} = \sum M_{\text{imp},i}$), respectively. The final training objective is:
\begin{equation}\label{eq:final_loss}
    \mathcal{L} = (1 - \lambda_{\text{SSIM}}) \mathcal{L}_{2} + \lambda_{\text{SSIM}} \mathcal{L}_{\text{D-SSIM}} + \lambda_{\text{POP}} \mathcal{L}_{\text{POP}}.
\end{equation}

In our implementation, we set $\lambda_{\text{SSIM}} = 0.01$ to balance the gradient magnitude of structural similarity with the pixel-wise intensity difference, as illustrated by the orange line in \cref{fig:grad_l1vsl2}. Furthermore, we apply a relatively small weight of $\lambda_{\text{POP}} = 0.001$ for the POP loss, which is sufficient to enforce opacity binarization.

Crucially, driving the opacities of important Gaussians toward $1$ more readily triggers Early Ray Termination. As shown in \cref{fig:perPixel_GSs_360_v2}, this mechanism enables POP to significantly reduce the number of Gaussians processed per pixel, yielding substantial gains in rendering efficiency.

\paragraph{Gradient Modulation and Leakage Mitigation.}
Beyond rendering efficiency, POP enhances the optimization of scene geometry by modulating the training signal. To understand how opacity affects densification, we analyze the view-space gradient of a Gaussian $\mathcal{G}_i$ for a pixel $\mathbf{p}$:
\begin{equation} \label{eq:ndc_grad}
    \frac{\partial \mathcal{L}}{\partial \boldsymbol{\mu}_{i}^{2D}} = 
    \left( \frac{\partial \mathcal{L}}{\partial \mathbf{c}(\mathbf{p})} \cdot \frac{\partial \mathbf{c}(\mathbf{p})}{\partial \alpha_i} \right) 
    \frac{\partial \alpha_i}{\partial \boldsymbol{\mu}_{i}^{2D}}.
\end{equation}
As shown in \cref{eq:loss_grads} and derived in the supplementary material, the terms $\frac{\partial \mathcal{L}}{\partial \mathbf{c}(\mathbf{p})}$ and $\frac{\partial \mathbf{c}(\mathbf{p})}{\partial \alpha_i}$ are independent of the opacity $o_i$. However, expanding the definition of $\alpha_i$ from \cref{eq:render}, we see that $\alpha_i \propto o_i$:
\begin{equation}\label{eq:alpha_i}
\alpha_i = o_i \exp \!\left( -\tfrac{1}{2} (\mathbf{p} - \boldsymbol{\mu}_i^{2D})^{\!T} (\boldsymbol{\Sigma}_i^{2D})^{-1} (\mathbf{p} - \boldsymbol{\mu}_i^{2D}) \right).
\end{equation}

Consequently, the first term in \cref{eq:opacity_loss_all} effectively amplifies the gradient feedback for important Gaussians, making them more sensitive to reconstruction errors and facilitating more effective densification in critical areas. Conversely, the second term drives uninformative Gaussians toward zero opacity. This not only attenuates their backpropagated gradients---thereby hindering them from triggering densification (\cref{eq:densification})---but also ensures they are naturally removed by the pruning mechanism. Through this dual mechanism of suppressing redundant densification and promoting pruning, our approach successfully mitigates gradient leakage. 

In particular, POP enables continuous and aggressive pruning without relying on heuristic opacity resets (\cref{subsec:preliminary}). By eliminating this mechanism, our approach maintains a smooth optimization trajectory, effectively reducing both memory footprint and training time (as detailed in \cref{subsec:ablation_opacity}).

% ===================================================================
% SECTION 4: Experiments
% ===================================================================
\section{Experiments}
\label{sec:experiments}

\subsection{Experiment Settings}
\label{subsec:experiment_settings}

\paragraph{Datasets and Metrics.}
We conduct experiments on three real-world datasets: all nine scenes from Mip-NeRF 360~\cite{Mip_NeRF_360} (four indoor and five outdoor), two outdoor scenes (\emph{train} and \emph{truck}) from Tanks and Temples~\cite{tanks}, and two indoor scenes (\emph{drjohnson} and \emph{playroom}) from Deep Blending~\cite{deep_blending}.

We evaluate rendering quality using Peak Signal-to-Noise Ratio (PSNR), Structural Similarity Index (SSIM), and Learned Perceptual Image Patch Similarity (LPIPS)~\cite{lpips}.
In addition, we report training time and the total number of Gaussians to assess model compactness and efficiency.

\paragraph{Implementation Details.}
Following the original 3DGS training schedule, we perform densification every 100 iterations from 0.5K to 15K steps.
After 15K iterations, the number of Gaussians remains fixed, and only their parameters are optimized.
Since the Polarized Opacity Prior (POP) primarily facilitates more effective densification, it is applied only during the first 15K iterations.
Furthermore, the \emph{blur-based densification} module is activated between 3K and 7K iterations. Additional details and ablation studies are provided in \cref{sec:ablation}.

\paragraph{Comparisons.}
We compare our method with the baseline 3DGS~\cite{3DGS}, two compression-oriented methods---PUP-3DGS~\cite{pup3dgs} and Mini-Splatting~\cite{miniSplatting}---and two fast-training methods---Taming-3DGS~\cite{Taming_3DGS} and Speedy-Splat~\cite{Speedy-Splat}.
Since our implementation is built upon the gsplat~\cite{gsplat} codebase, we also include gsplat in our comparisons.
For a fair evaluation, most methods are retrained and tested on the same NVIDIA RTX 4090 GPU.

\input{tables/main_eval_360_and_deep}
\input{tables/main_eval_tank}

\subsection{Experimental Results}
\label{subsec:experimental_results}

The quantitative results are shown in \cref{tab:360_deep_combined_horiz} and \cref{tab:eval_tanks_and_temples}. The visual results are also shown in \cref{fig:qualitative_comparison}.
For the compression-based baselines, PUP-3DGS and Mini-Splatting, we initially follow the default configurations from their original papers.
Specifically, PUP-3DGS defaults to compressing each model to one-tenth of its original size.

On the Mip-NeRF 360 and Deep Blending datasets (\cref{tab:360_deep_combined_horiz}), our method achieves the fastest training time and the lowest Gaussian count among all compared methods. Notably, on the Mip-NeRF 360 dataset, our approach surpasses both PUP-3DGS ($312$K) and Speedy-Splat ($279$K) in PSNR with only $268$K Gaussians.

On the Tanks and Temples dataset (\cref{tab:eval_tanks_and_temples}), our method achieves the third-best PSNR score by using only $\approx$270K Gaussians. Notably, we attain the fastest training speed among all compared methods, reducing training time by at least 30\% relative to the next fastest state-of-the-art approach, Taming-3DGS. 
We note that the default settings for PUP-3DGS and Mini-Splatting result in far more aggressive compression than our method.
For a fair comparison of visual quality, we also report results (marked with ${*}$) where we adjusted their hyperparameters to produce a Gaussian count comparable to ours.
Specifically, we modified the PUP-3DGS schedule from $100 \rightarrow 20\% \rightarrow 10\%$ to $100 \rightarrow 30\% \rightarrow 15\%$ and increased the initial Mini-Splatting sampling factor from $0.3$ to $0.4$.

In summary, our method demonstrates a superior trade-off between efficiency and quality. As shown in \cref{fig:teaser}, we achieve substantial reductions in training time and model size, yielding an intrinsically compact representation while maintaining comparable visual quality.

\input{tables/results_visualize}

% ===================================================================
% SECTION 5: Ablation Study
% ===================================================================
\section{Ablation Study}
\label{sec:ablation}

In this section, we evaluate the effectiveness of the key components in our model: the L2 loss, blur-based densification, and Polarized Opacity Prior (POP).

\subsection{$\mathcal{L}_2$ Loss and Blur-Based Densification}
\label{subsec:ablation_l2}

\begin{figure}[tb]
    \centering
    % 子圖 (a)：Gaussian Count
    \begin{subfigure}{0.48\linewidth}
        \centering
        \includegraphics[width=\linewidth]{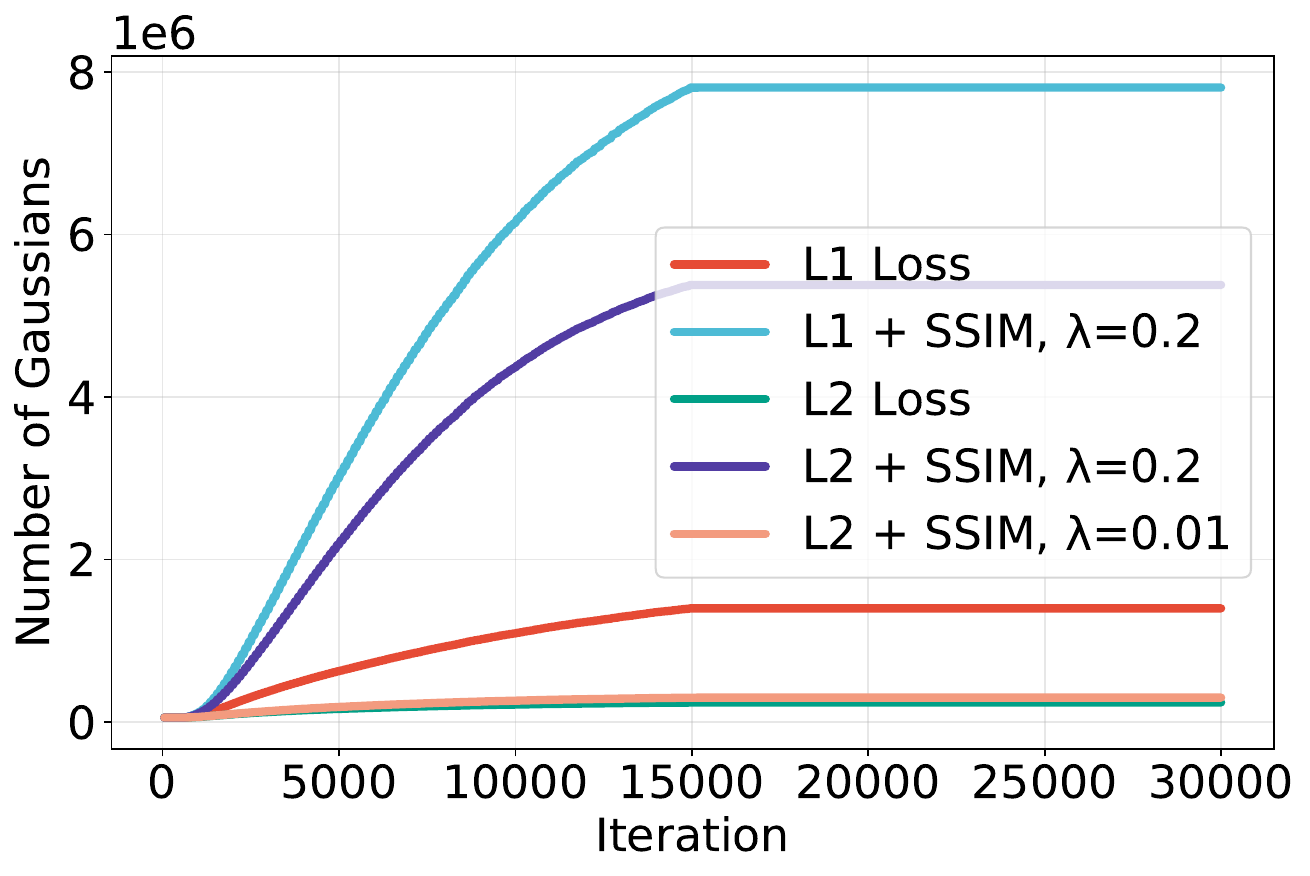}
        \caption{Gaussian counts during training (green curve lies consistently below the orange curve.)}
        \label{fig:gs_count}
    \end{subfigure}
    \hfill
    % 子圖 (b)：L1 vs L2 PSNR
    \begin{subfigure}{0.48\linewidth}
        \centering
        \includegraphics[width=\linewidth]{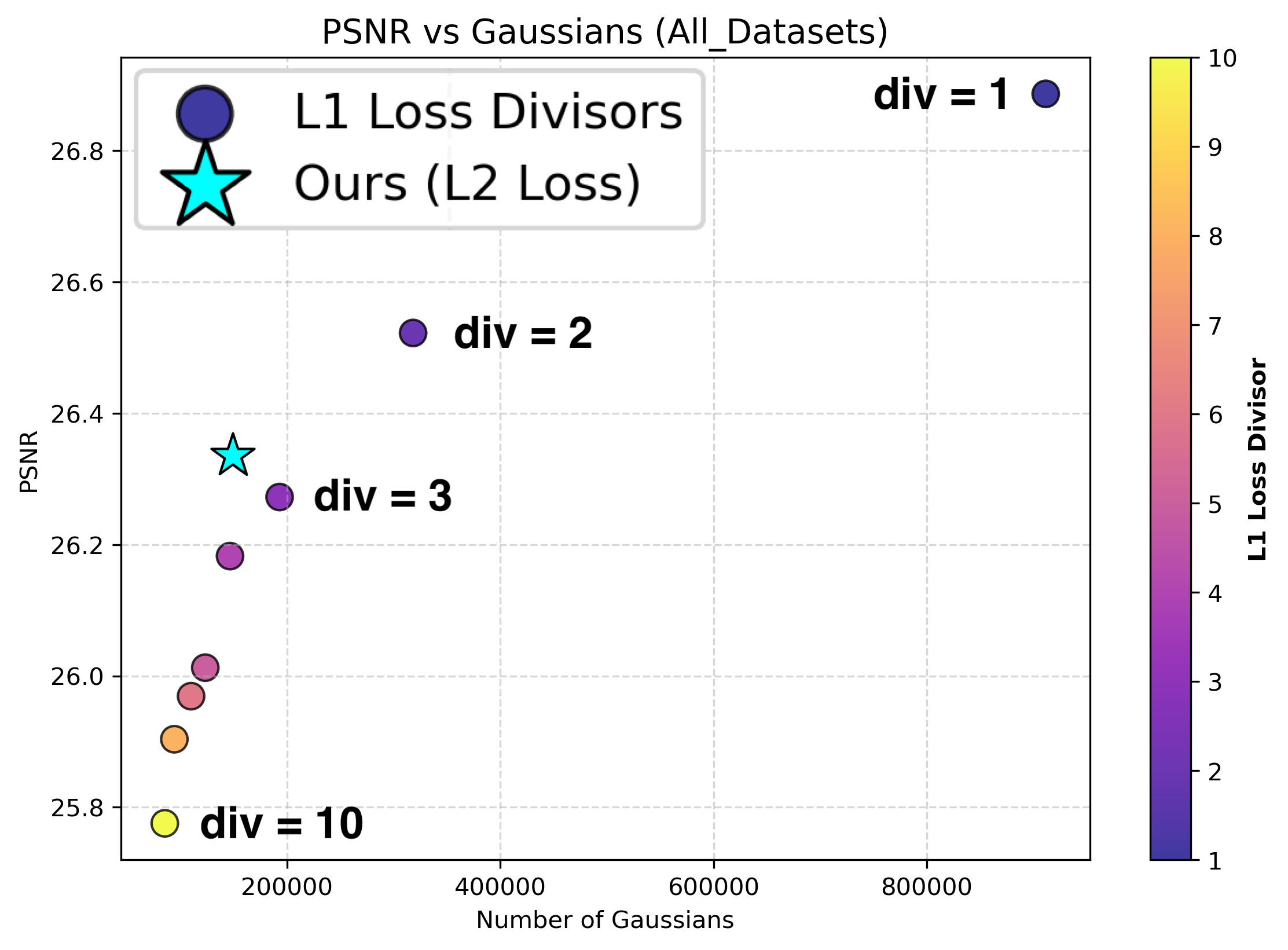}
        \caption{PSNR vs.\ Gaussian count}
        \label{fig:L1_vs_L2_PSNR}
    \end{subfigure}
    
    % 總標題：消除重複，專注解釋趨勢與意義
    \caption{Ablation on loss functions. (a) Impact of different supervisions on Gaussian count. (b) Trade-off between PSNR and Gaussian count for $\mathcal{L}_1$ with varying scaling factors (circles) and the $\mathcal{L}_2$ loss (star).}
    \label{fig:ablation_loss_functions}
\end{figure}

As discussed in \cref{subsec:l2_loss}, replacing the $\mathcal{L}_1$ loss with an $\mathcal{L}_2$ loss substantially reduces the overall gradient magnitude during the optimization process. This reduction directly translates to a more controlled densification process, resulting in a lower number of Gaussians, as illustrated in \cref{fig:gs_count}.

Unlike the constant $\frac{\partial \mathcal{L}_1}{\partial \mathbf{c}}$ gradient, $\frac{\partial \mathcal{L}_2}{\partial \mathbf{c}}$ scales proportionally with the reconstruction error, providing more informative optimization signals. As illustrated in \cref{fig:L1_vs_L2_PSNR}, we compare $\mathcal{L}_2$ against $\mathcal{L}_1$ with various scaling factors (denoted as $div=n$ for $\mathcal{L}_1/n$). While simply reducing the $\mathcal{L}_1$ gradient magnitude also lowers the Gaussian count, the $\mathcal{L}_2$ loss achieves a higher PSNR at a comparable number of Gaussians. This adaptive behavior effectively suppresses artifacts and yields superior training outcomes. Given its enhanced stability and efficiency, we adopt the $\mathcal{L}_2$ + SSIM loss ($\lambda_{\text{SSIM}} = 0.01$) as our baseline for all subsequent ablations.

\paragraph{Blur-Based Densification.}
\label{subsec:ablation_blur}
As shown in \cref{tab:ablation_all}, relying solely on $\mathcal{L}_2$+SSIM leads to a sparse Gaussian distribution, which limits the model's representational capacity. Integrating blur-based densification effectively restores the necessary Gaussian density (\cref{tab:ablation_all}, rows 2 and 4). This restoration process is evident in \cref{fig:total_GS_count_360_v2}, where our model's Gaussian count (orange line) rapidly increases between $3$k and $7$k iterations.

\cref{tab:blur_densification_all} further analyzes the impact of activating blur-based densification across different training stages. We initiate this process at $3$k iterations to ensure the model has reached a stable initialization. While extended schedules (e.g., $3$k--$15$k) consume a $2$--$3\times$ Gaussian budget and prolong training time, they yield inconsistent, dataset-dependent PSNR changes ($+0.16$ on Mip-NeRF 360, $-0.07$ on T\&T, $-0.37$ on DB). On the other hand, halting too early ($3$k--$4$k) results in under-densification. Therefore, we select the \textbf{$3$k--$7$k} schedule (highlighted in \textbf{gray}), which provides the optimal trade-off: ensuring sufficient density for accurate reconstruction while avoiding unnecessary model bloat and maintaining computational efficiency.

\input{tables/ablation_module}

\subsection{Polarized Opacity Prior (POP)}
\label{subsec:ablation_opacity}

\begin{figure}[tb]
    \centering
    \includegraphics[width=0.6\linewidth]{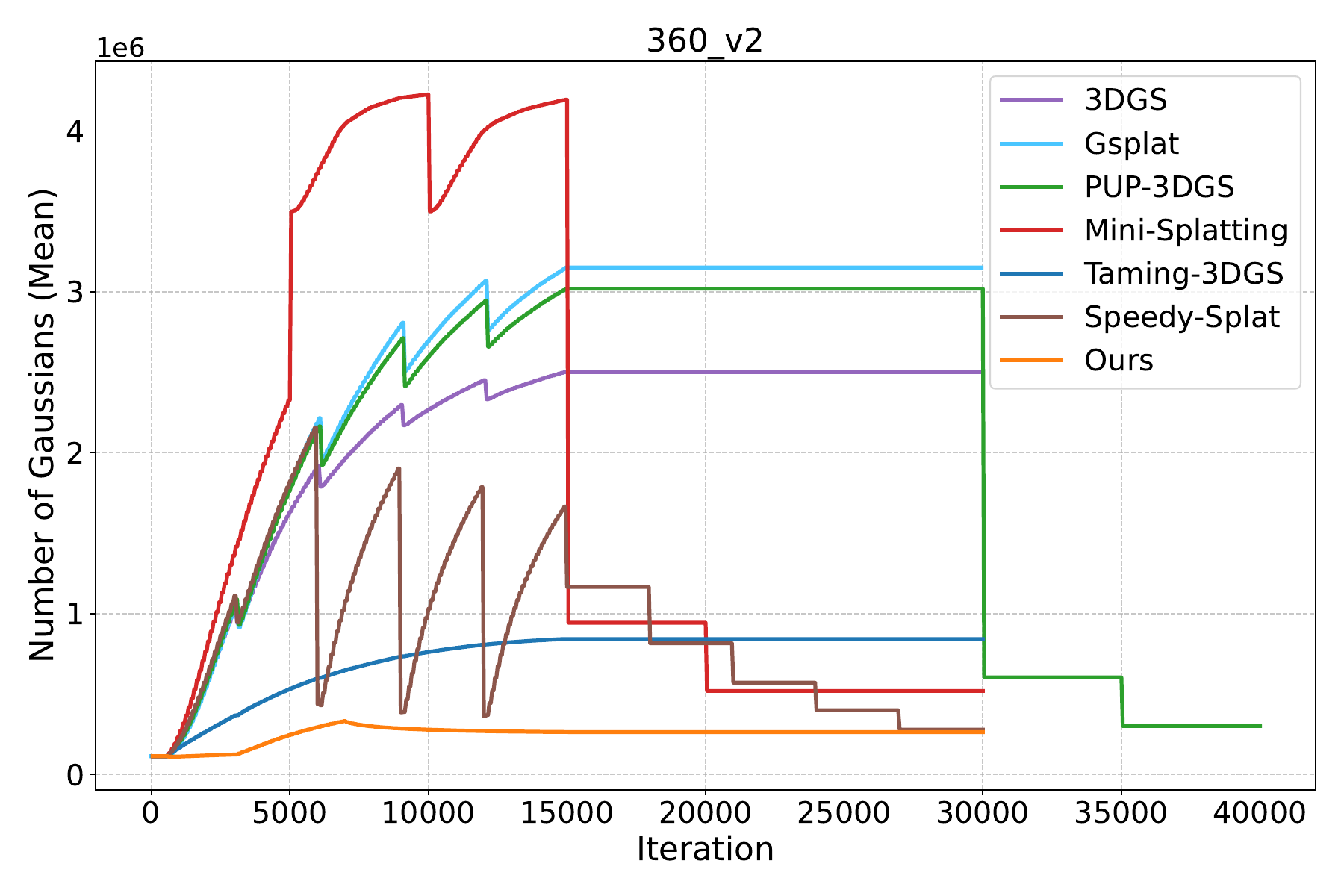}
    \caption{Average number of Gaussians during training on the Mip-NeRF 360~\cite{Mip_NeRF_360} dataset. Note that PUP-3DGS~\cite{pup3dgs} performs post-training compression after 30,000 iterations.}
    \label{fig:total_GS_count_360_v2}
\end{figure}

Periodic opacity resets inevitably induce sudden performance drops and severe disruptions during training. As shown in \cref{fig:total_GS_count_360_v2}, models relying on this strategy (e.g., 3DGS~\cite{3DGS}, gsplat~\cite{gsplat}, and PUP-3DGS~\cite{pup3dgs}) exhibit drastic fluctuations every 3K iterations during the first 15K steps. This forces the optimization process to spend thousands of iterations recovering stability with noisy, unreliable gradients. In contrast, POP introduces an opacity regularization term (\cref{eq:opacity_loss_all}) that continuously and selectively suppresses redundant Gaussians while preserving essential ones. This mechanism eliminates such instability, maintaining a smooth training trajectory (orange curve). Furthermore, the results in \cref{tab:ablation_pop_final} demonstrate that this smooth optimization trajectory allows POP to achieve comparable rendering quality using fewer Gaussians and reduced training time, thereby significantly improving overall training efficiency.

Notably, \cref{fig:total_GS_count_360_v2} shows that while Taming-3DGS~\cite{Taming_3DGS} also exhibits gradual Gaussian growth, it relies on a complex Parabolic Schedule and Score-based Sampling for targeted densification. In contrast, our approach seamlessly synergizes with the native pruning mechanism to achieve organically controlled growth throughout training. This simplicity underscores the robustness and effectiveness of the POP design.

\input{tables/ablation_blur_dense_iter}
\input{tables/ablation_opacity}

% ===================================================================
% SECTION 6: Conclusion
% ===================================================================
\section{Limitations}
\label{sec:limitation}

Despite its favorable speed-compactness trade-off, our method has several limitations. 
First, the blur-based densification relies on a fixed schedule (3k--7k iterations), which may not generalize to scenes with diverse geometric complexities. 

Second, while achieving high compactness, our model occasionally struggles with high-frequency details in complex textures. Due to the significantly reduced Gaussian count, certain artifacts or rendering gaps may become more apparent compared to the original 3DGS~\cite{3DGS} (detailed visual comparisons are 
provided in the supplementary material).

Finally, despite its compactness, our model does not yet employ post-processing compression techniques---such as vector quantization or codebook-based encoding---which could further optimize storage efficiency.

\section{Conclusion}
\label{sec:conclusion}
In this paper, we presented an efficient framework for compact 3DGS scene representation, offering a streamlined alternative to the redundant ``densify-then-prune'' paradigm. By integrating $\mathcal{L}_2$ loss, targeted blur-based densification, and a novel polarized opacity prior term, our method constructs a compact model from the outset---suppressing uninformative Gaussians early while selectively refining essential structures. 
The proposed approach significantly accelerates training and reduces model size while maintaining comparable rendering quality. Unlike techniques that rely on complex MLPs or auxiliary attributes, our design remains lightweight by adhering to the original 3DGS formulation, ensuring low architectural overhead and fast convergence. We believe this simple yet effective framework serves as a practical and robust baseline for future research in efficient 3DGS reconstruction.

\section*{Acknowledgements}
This work was supported in part by the National Science and Technology Council, Taiwan, under grants NSTC 115-2634-F-007-003 and 113-2221-E-007-104-MY3.

\bibliographystyle{splncs04}
\bibliography{main}

\end{document}

%% file: tables/main_eval_360_and_deep.tex
\begin{table*}[t]
    \centering
    
    \caption{Quantitative evaluation on \textbf{Mip-NeRF 360} and \textbf{Deep Blending}. Training time and checkpoint size (Ckpt) are reported in \textbf{seconds} and \textbf{MB}, respectively. Best, second-best, and third-best results are highlighted in \textcolor{bestred}{red}, \textcolor{secondorange}{orange}, and \textcolor{thirdyellow}{yellow}, respectively.}
    \label{tab:360_deep_combined_horiz}
    
    \renewcommand{\tabcolsep}{1.8pt}
    {\footnotesize
    \resizebox{\textwidth}{!}{
    \begin{tabular}{@{}l|cccccc|cccccc@{}}
    \toprule
    \textbf{Dataset} & \multicolumn{6}{c|}{\textbf{Mip-NeRF 360}~\cite{Mip_NeRF_360}} & \multicolumn{6}{c}{\textbf{Deep Blending}~\cite{deep_blending}} \\
    \textbf{Method} & Num $\downarrow$ & Time $\downarrow$ & Ckpt $\downarrow$ & PSNR $\uparrow$ & SSIM $\uparrow$ & LPIPS $\downarrow$ & Num $\downarrow$ & Time $\downarrow$ & Ckpt $\downarrow$ & PSNR $\uparrow$ & SSIM $\uparrow$ & LPIPS $\downarrow$ \\
    \midrule

    3DGS~\cite{3DGS} 
    & 2,509k & 1,304 & 595 & \cellcolor{thirdyellow}27.250 & \cellcolor{thirdyellow}0.811 & \cellcolor{thirdyellow}0.226 
    & 2,484k & 1,197 & 588 & \cellcolor{thirdyellow}29.847 & \cellcolor{secondorange}0.907 & \cellcolor{bestred}0.089 \\

    Gsplat~\cite{gsplat} 
    & 3,168k & 886 & 713 & \cellcolor{bestred}27.661 & \cellcolor{secondorange}0.824 & \cellcolor{bestred}0.166 
    & 3,042k & 789 & 685 & 29.573 & 0.904 & 0.169 \\

    PUP-3DGS~\cite{pup3dgs} 
    & \cellcolor{thirdyellow}312k & 1,216 & \cellcolor{thirdyellow}74 & 26.772 & 0.795 & 0.258 
    & \cellcolor{thirdyellow}282k & 1,205 & \cellcolor{thirdyellow}67 & 29.471 & 0.903 & \cellcolor{secondorange}0.103 \\

    Mini-Splatting~\cite{miniSplatting} 
    & 495k & 896 & 117 & \cellcolor{secondorange}27.605 & \cellcolor{bestred}0.833 & \cellcolor{secondorange}0.202 
    & 348k & 834 & 82 & \cellcolor{bestred}29.987 & \cellcolor{bestred}0.908 & 0.253 \\

    Taming-3DGS~\cite{Taming_3DGS} 
    & 346k & \cellcolor{secondorange}329 & 82 & 27.147 & 0.772 & 0.294 
    & 294k & \cellcolor{secondorange}282 & 69 & \cellcolor{secondorange}29.886 & \cellcolor{thirdyellow}0.905 & 0.270 \\

    Speedy-Splat~\cite{Speedy-Splat} 
    & \cellcolor{secondorange}279k & \cellcolor{thirdyellow}728 & \cellcolor{secondorange}66 & 26.926 & 0.783 & 0.294 
    & \cellcolor{secondorange}250k & \cellcolor{thirdyellow}667 & \cellcolor{secondorange}59 & 29.632 & 0.904 & \cellcolor{thirdyellow}0.108 \\

    \hline
    \rowcolor[gray]{0.95} Ours 
    & \cellcolor{bestred}268k & \cellcolor{bestred}279 & \cellcolor{bestred}60 & 27.013 & 0.760 & 0.277 
    & \cellcolor{bestred}118k & \cellcolor{bestred}256 & \cellcolor{bestred}27 & 28.626 & 0.869 & 0.266 \\
    
    \bottomrule
    \end{tabular}
    }
    }
\end{table*}

%% file: tables/main_eval_tank.tex
\begin{table*}[t]
    \centering
    
    \caption{Quantitative evaluation on \textbf{Tanks \& Temples}. $\ast$ indicates that the method is evaluated under a reduced compression strength for fair comparison.}
    \label{tab:eval_tanks_and_temples}
    
    \renewcommand{\arraystretch}{0.9} % 讓行間距變緊湊（扁一點）
    \renewcommand{\tabcolsep}{3pt}    % 欄位水平間距
    \begin{tabular}{@{}l@{\,\,}|rccccc@{}}
    \toprule
    Dataset & \multicolumn{6}{c}{Tanks \& Temples} \\
    Method & Num $\downarrow$ & Time (s) $\downarrow$ & Ckpt (MB) $\downarrow$ & PSNR $\uparrow$ & SSIM $\uparrow$ & LPIPS $\downarrow$ \\
    \midrule
    3DGS ~\cite{3DGS}  & 1,573k & 683 & 372 & \cellcolor{bestred}23.808 & \cellcolor{bestred}0.853 & \cellcolor{bestred}0.091 \\
    Gsplat ~\cite{gsplat}  & 1,840k & 543 & 414 & 23.508 & \cellcolor{secondorange}0.845 & \cellcolor{secondorange}0.127 \\
    PUP-3DGS ~\cite{pup3dgs} & \cellcolor{secondorange}183k & 745 & \cellcolor{bestred}43 & 22.672 & 0.804 & 0.161 \\
    PUP-3DGS$^\ast$ ~\cite{pup3dgs} & 275k & 766 & 65 & 23.049 & 0.818 & 0.222 \\
    Mini-Splatting ~\cite{miniSplatting}  & \cellcolor{thirdyellow}201k & 591 & \cellcolor{thirdyellow}48 & 23.212 & 0.835 & 0.202 \\
    Mini-Splatting$^\ast$ ~\cite{miniSplatting} & 254k & 613 & 61 & 23.339 & \cellcolor{thirdyellow}0.842 & 0.189 \\
    Taming-3DGS ~\cite{Taming_3DGS}  & 318k & \cellcolor{secondorange}305 & 75 & \cellcolor{secondorange}23.714 & 0.834 & 0.211 \\
    Speedy-Splat ~\cite{Speedy-Splat}  & \cellcolor{bestred}182k & \cellcolor{thirdyellow}404 & \cellcolor{bestred}43 & 23.403 & 0.819 & \cellcolor{thirdyellow}0.142 \\
    \hline
    Ours & 269k & \cellcolor{bestred}208 & 60 & \cellcolor{thirdyellow}23.618 & 0.811 & 0.181 \\
    \bottomrule
    \end{tabular}
\end{table*}

%% file: tables/results_visualize.tex
% \begin{figure*}[htbp]
\begin{figure*}[tb]  % 使用 ! 忽略限制，並強制置頂
    \centering
    \setlength{\tabcolsep}{3pt} % Reduce horizontal spacing between columns
    \begin{tabular}{lcccc}
        \toprule % Top rule from booktabs
        & \textbf{Truck} & \textbf{Train} & \textbf{Playroom} & \textbf{Bonsai} \\
        \midrule % Mid rule from booktabs
        
        \raisebox{21pt}{\rotatebox{90}{\textbf{Ours}}} & 
            \includegraphics[width=0.22\linewidth,keepaspectratio]{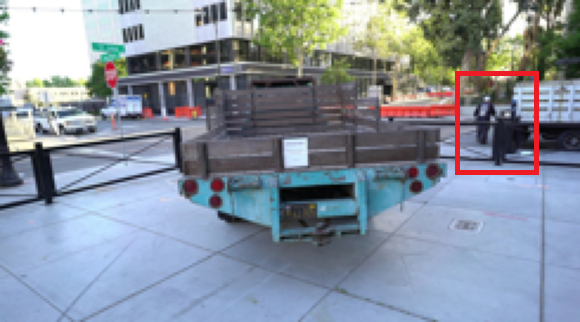} & 
            \includegraphics[width=0.22\linewidth,keepaspectratio]{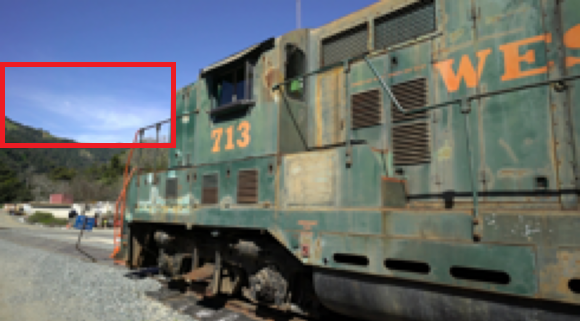} & 
            \includegraphics[width=0.22\linewidth,keepaspectratio]{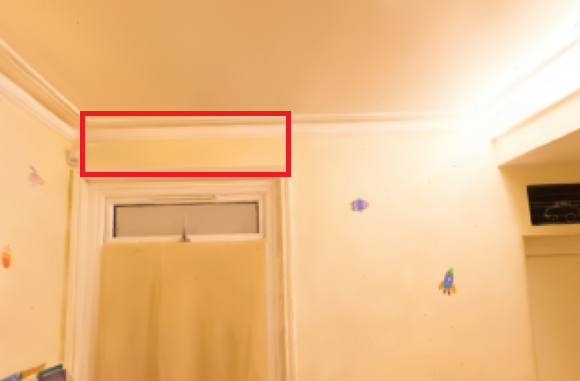} & 
            \includegraphics[width=0.22\linewidth,keepaspectratio]{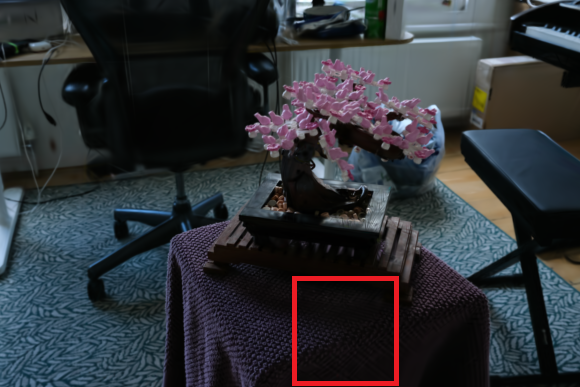} \\
        
        \raisebox{21pt}{\rotatebox{90}{\textbf{3DGS}}} & 
            \includegraphics[width=0.22\linewidth,keepaspectratio]{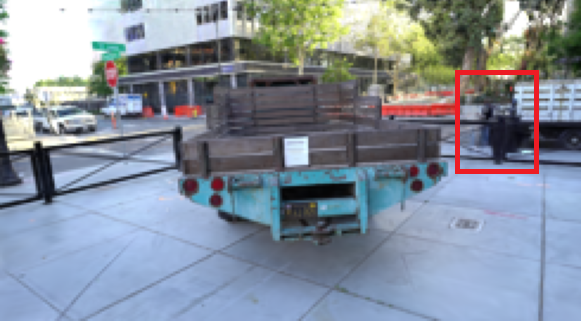} & 
            \includegraphics[width=0.22\linewidth,keepaspectratio]{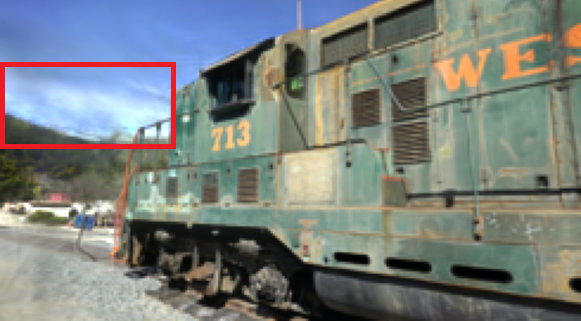} & 
            \includegraphics[width=0.22\linewidth,keepaspectratio]{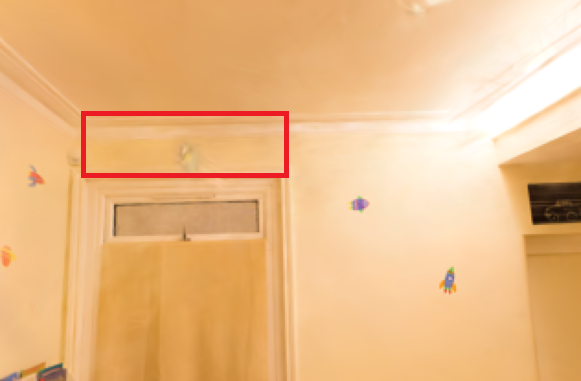} & 
            \includegraphics[width=0.22\linewidth,keepaspectratio]{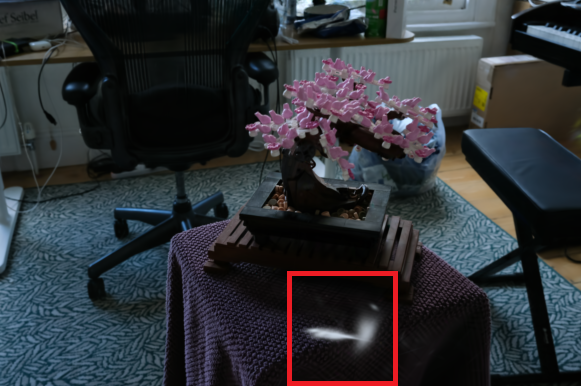} \\
        
        \raisebox{21pt}{\rotatebox{90}{\textbf{GT}}} & 
            \includegraphics[width=0.22\linewidth,keepaspectratio]{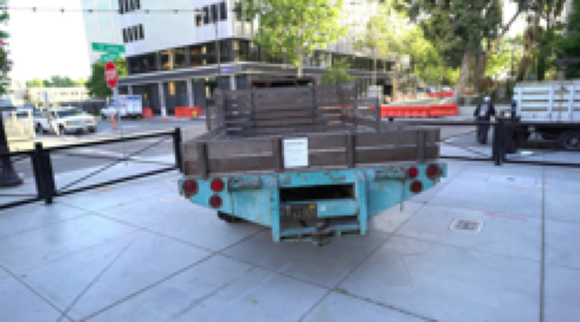} & 
            \includegraphics[width=0.22\linewidth,keepaspectratio]{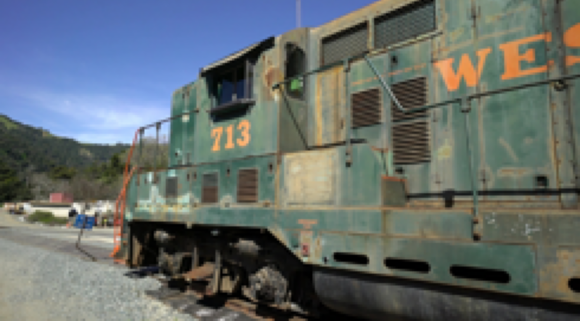} & 
            \includegraphics[width=0.22\linewidth,keepaspectratio]{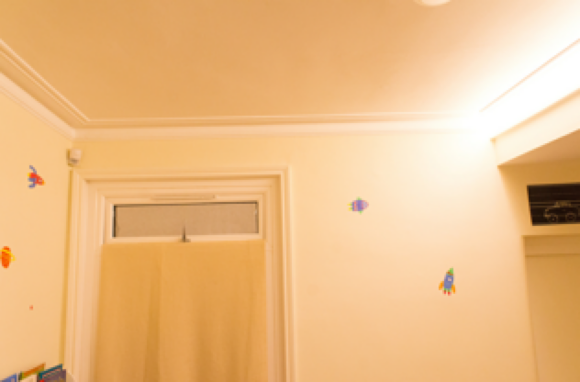} & 
            \includegraphics[width=0.22\linewidth,keepaspectratio]{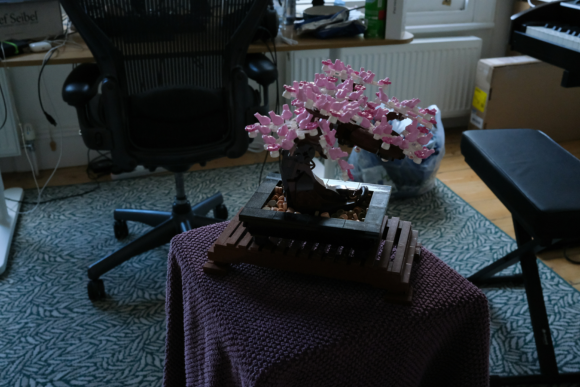} \\
        
        \bottomrule % Bottom rule from booktabs
    \end{tabular}
    \caption{Qualitative comparison on various scenes. Our method (top) achieves higher reconstruction fidelity with fewer artifacts compared to 3DGS (middle). Red boxes highlight improved fine details (e.g., Truck, Bonsai) and smoother surface reconstruction (e.g., Train, Playroom).}
    \label{fig:qualitative_comparison}
\end{figure*}

%% file: tables/ablation_module.tex
\begin{table*}[t]
    \centering
    \caption{\textbf{Ablation on our components}. ``Blur'' denotes our Blur-based Densification module. Our full method is highlighted in \colorbox[gray]{0.95}{gray}. \textbf{Bold} numbers indicate the best performance for each metric.}
    \label{tab:ablation_all}
    \renewcommand{\tabcolsep}{2.5pt}
    {\footnotesize
    \resizebox{\textwidth}{!}{
    \begin{tabular}{@{}ccc|ccccc|ccccc|ccccc@{}}
    \toprule
    \multicolumn{3}{c|}{\textbf{Modules}} & \multicolumn{5}{c|}{\textbf{Mip-NeRF 360}} & \multicolumn{5}{c|}{\textbf{Tanks \& Temples}} & \multicolumn{5}{c}{\textbf{Deep Blending}} \\
    L2 & Blur & POP & Num $\downarrow$ & Time $\downarrow$ & PSNR $\uparrow$ & SSIM $\uparrow$ & LPIPS $\downarrow$ & Num $\downarrow$ & Time $\downarrow$ & PSNR $\uparrow$ & SSIM $\uparrow$ & LPIPS $\downarrow$ & Num $\downarrow$ & Time $\downarrow$ & PSNR $\uparrow$ & SSIM $\uparrow$ & LPIPS $\downarrow$ \\
    \midrule
    \cmark & \xmark & \xmark & 210k & 271 & 26.57 & 0.725 & 0.339 & 221k & 271 & 23.53 & 0.809 & 0.191 & 193k & 259 & 28.51 & 0.870 & 0.252 \\
    \cmark & \cmark & \xmark & 583k & 343 & \textbf{27.22} & \textbf{0.771} & \textbf{0.255} & 557k & 267 & \textbf{23.71} & \textbf{0.821} & \textbf{0.161} & 297k & 275 & 28.56 & \textbf{0.873} & \textbf{0.243} \\
    \cmark & \xmark & \cmark & \textbf{140k} & \textbf{264} & 26.28 & 0.715 & 0.354 & \textbf{157k} & \textbf{176} & 23.46 & 0.798 & 0.209 & \textbf{94k} & \textbf{254} & 28.43 & 0.867 & 0.272 \\
    \hline
    \rowcolor[gray]{0.95} 
    \cmark & \cmark & \cmark & 268k & 279 & 27.01 & 0.760 & 0.277 & 269k & 209 & 23.62 & 0.811 & 0.181 & 118k & 257 & \textbf{28.63} & 0.869 & 0.266 \\
    \bottomrule
    \end{tabular}
    } % End resizebox
    } % End footnotesize
\end{table*}

%% file: tables/ablation_blur_dense_iter.tex
\begin{table*}[t]
    \centering

    \caption{Impact of blur-based densification schedules. We highlight our balanced 3k--7k setting in \colorbox[gray]{0.95}{gray}. \textbf{Bold} numbers indicate the best performance for each metric.}
    \label{tab:blur_densification_all}
    
    \renewcommand{\tabcolsep}{2pt}
    {\footnotesize
    \resizebox{\textwidth}{!}{
    \begin{tabular}{@{}l|ccccc|ccccc|ccccc@{}}
    \toprule
    \textbf{Dataset} & \multicolumn{5}{c|}{\textbf{Mip-NeRF 360}} & \multicolumn{5}{c|}{\textbf{Tanks \& Temples}} & \multicolumn{5}{c}{\textbf{Deep Blending}} \\
    \textbf{Blur Iter.} & Num $\downarrow$ & Time $\downarrow$ & PSNR $\uparrow$ & SSIM $\uparrow$ & LPIPS $\downarrow$ & Num $\downarrow$ & Time $\downarrow$ & PSNR $\uparrow$ & SSIM $\uparrow$ & LPIPS $\downarrow$ & Num $\downarrow$ & Time $\downarrow$ & PSNR $\uparrow$ & SSIM $\uparrow$ & LPIPS $\downarrow$ \\
    \midrule
    3k--4k & \textbf{175k} & \textbf{269} & 26.54 & 0.736 & 0.321 & \textbf{185k} & \textbf{186} & 23.45 & 0.804 & 0.196 & \textbf{100k} & \textbf{257} & 28.21 & 0.863 & 0.271 \\
    
    \rowcolor[gray]{0.95} 
    \textbf{3k--7k} & 268k & 279 & 27.01 & 0.760 & 0.277 & 269k & 209 & \textbf{23.62} & 0.811 & 0.181 & 118k & \textbf{257} & \textbf{28.63} & \textbf{0.869} & 0.266 \\
    
    3k--11k & 394k & 326 & 27.13 & 0.768 & 0.260 & 412k & 238 & 23.55 & 0.814 & 0.172 & 188k & 264 & 28.53 & \textbf{0.869} & \textbf{0.261} \\
    3k--15k & 562k & 358 & \textbf{27.17} & \textbf{0.773} & \textbf{0.250} & 647k & 275 & 23.55 & \textbf{0.816} & \textbf{0.165} & 331k & 284 & 28.26 & 0.867 & 0.262 \\
    \bottomrule
    \end{tabular}
    } % End resizebox
    } % End footnotesize
\end{table*}

%% file: tables/ablation_opacity.tex
\begin{table*}[t]
    \centering
    \caption{Ablation study of our \textbf{Polarized Opacity Prior (POP)} against the standard \textbf{Opacity Reset}. POP allows for a more compact representation and efficient training while maintaining competitive visual quality across all datasets.}
    \label{tab:ablation_pop_final}
    \setlength{\tabcolsep}{0.5pt} % 縮小欄位間距
    \resizebox{\textwidth}{!}{ % 強制縮放至版面寬度
    \begin{tabular}{@{}l|ccccc|ccccc|ccccc@{}}
    \toprule
    \textbf{Dataset} & \multicolumn{5}{c|}{\textbf{Mip-NeRF 360}} & \multicolumn{5}{c|}{\textbf{Tanks \& Temples}} & \multicolumn{5}{c}{\textbf{Deep Blending}} \\
    \textbf{Method} & Num$\downarrow$ & Time$\downarrow$ & PSNR$\uparrow$ & SSIM$\uparrow$ & LPIPS$\downarrow$ & Num$\downarrow$ & Time$\downarrow$ & PSNR$\uparrow$ & SSIM$\uparrow$ & LPIPS$\downarrow$ & Num$\downarrow$ & Time$\downarrow$ & PSNR$\uparrow$ & SSIM$\uparrow$ & LPIPS$\downarrow$ \\
    \midrule
    Reset & 548k & 328 & 26.82 & \textbf{0.780} & \textbf{0.260} & 432k & 247 & 22.91 & 0.795 & 0.201 & 234k & 270 & 28.51 & \textbf{0.871} & \textbf{0.258} \\
    \rowcolor[gray]{0.95} POP & \textbf{268k} & \textbf{279} & \textbf{27.01} & 0.760 & 0.277 & \textbf{269k} & \textbf{209} & \textbf{23.62} & \textbf{0.811} & \textbf{0.181} & \textbf{118k} & \textbf{257} & \textbf{28.63} & 0.869 & 0.266 \\
    \bottomrule
    \end{tabular}
    }
\end{table*}